\documentclass[10pt,twocolumn,letterpaper]{article}

\usepackage{cvpr}              

\usepackage{graphicx}
\usepackage{amsmath}
\usepackage{amssymb}
\usepackage{booktabs}
\usepackage[accsupp]{axessibility}

\usepackage{xcolor}
\definecolor{citecolor}{HTML}{0071bc}

\newlength\savewidth\newcommand\shline{\noalign{\global\savewidth\arrayrulewidth
  \global\arrayrulewidth 1pt}\hline\noalign{\global\arrayrulewidth\savewidth}}
  
  \makeatletter 
 \def\@fnsymbol#1{\ensuremath{\ifcase#1\or \dagger\or \ddagger\or
   \mathsection\or \mathparagraph\or \|\or **\or \dagger\dagger
   \or \ddagger\ddagger \else\@ctrerr\fi}}
\makeatother

\definecolor{mypink}{RGB}{10,186,181}

\usepackage[pagebackref,breaklinks,colorlinks]{hyperref}

\definecolor{citecolor}{HTML}{0071bc}
\definecolor{paleplum}{rgb}{0.8, 0.6, 0.8}
\hypersetup{
  colorlinks,
  citecolor=citecolor,
  linkcolor=red
}

\usepackage[pagebackref,breaklinks,colorlinks]{hyperref}
\usepackage{multirow}

\usepackage[capitalize]{cleveref}
\crefname{section}{Sec.}{Secs.}
\Crefname{section}{Section}{Sections}
\Crefname{table}{Table}{Tables}
\crefname{table}{Tab.}{Tabs.}

\def\cvprPaperID{4878} 
\def\confName{CVPR}
\def\confYear{2023}

\begin{document}

\title{Dynamic Focus-aware Positional Queries for Semantic Segmentation}

\author{%
  Haoyu He$^{1}$ ~~ Jianfei Cai$^{1}$ ~~ Zizheng Pan$^{1}$ ~~ Jing Liu$^{1}$ ~~ \\  Jing Zhang$^{2}$ ~~ Dacheng Tao$^{2}$ ~~ Bohan Zhuang$^{1}\thanks{Corresponding author. E-mail: $\tt  bohan.zhuang@gmail.com$}$ \\ [0.25cm]
$^1$ ZIP Lab, Monash University \quad $^2$ The University of Sydney \\[0.1cm]}
\maketitle

\begin{abstract}
The DETR-like segmentors have underpinned the most recent breakthroughs in semantic segmentation, which end-to-end train a set of queries representing the class prototypes or target segments. Recently, masked attention~\cite{cheng2021masked} is proposed to restrict each query to only attend to the foreground regions predicted by the preceding decoder block for easier optimization. Although promising, it relies on the learnable parameterized positional queries which tend to encode the dataset statistics, leading to inaccurate localization for distinct individual queries. In this paper, we propose a simple yet effective query design for semantic segmentation termed Dynamic Focus-aware Positional Queries (DFPQ), which dynamically generates positional queries conditioned on the cross-attention scores from the preceding decoder block and the positional encodings for the corresponding image features, simultaneously. Therefore, our DFPQ preserves rich localization information for the target segments and provides accurate and fine-grained positional priors. In addition, we propose to efficiently deal with high-resolution cross-attention by only aggregating the contextual tokens based on the low-resolution cross-attention scores to perform local relation aggregation. Extensive experiments on ADE20K and Cityscapes show that with the two modifications on Mask2former, our framework achieves SOTA performance and outperforms Mask2former by clear margins of 1.1\%, 1.9\%, and 1.1\% single-scale mIoU with ResNet-50, Swin-T, and Swin-B backbones on the ADE20K validation set, respectively. Source code is available at \url{https://github.com/ziplab/FASeg}.
\end{abstract}

\section{Introduction}

Semantic segmentation aims at assigning each pixel in an image with a semantic class label. As the end-to-end Detection Transfomer (DETR)~\cite{carion2020end,zhu2021deformable,yao2021efficient,wang2022towards} is revolutionizing the paradigm of the object detection task, recent segmentors~\cite{zhang2021k,bousselham2021efficient,cheng2021per,cheng2021masked} follow DETR to learn a set of queries representing the class prototypes or target segments and achieve state-of-the-art performance on semantic segmentation.

\begin{figure*}[t!]
  \centering
  \includegraphics[width=\linewidth]{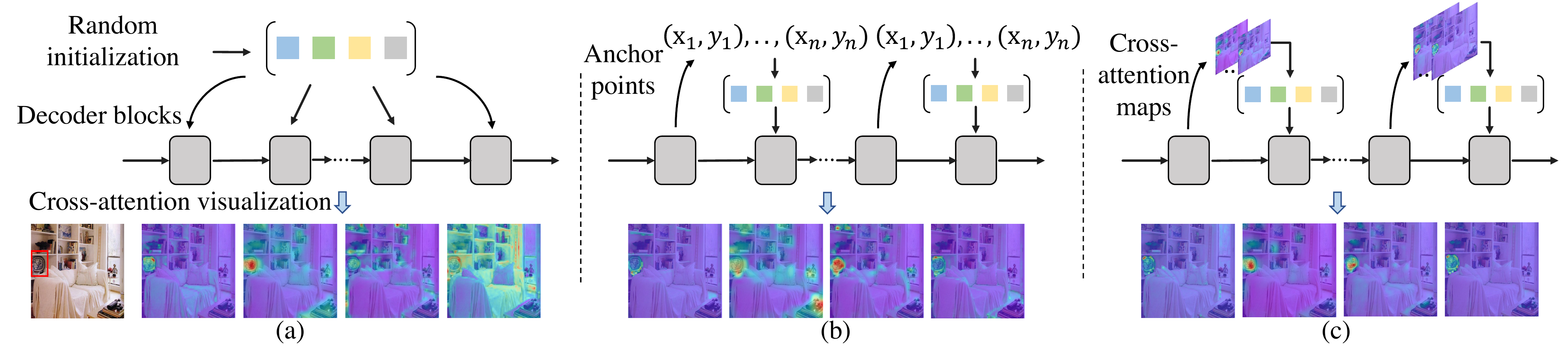}
  \vspace{-2.em}
  \caption{(a) The original randomly initialized positional queries~\cite{carion2020end} as learnable network parameters, where the positional queries are shared among the Transformer decoder blocks and tend to encode dataset statistics modelling the likely positions for the semantic regions, which leads to inaccurate localization. (b) The anchor-based positional queries~\cite{wang2021anchor} are conditional on the bounding box coordinates to give each query positional priors around the anchor. However, the anchor points cannot describe semantic regions, thus still sub-optimal for semantic segmentation. (c) Our dynamic focus-aware queries for semantic segmentation are dynamically generated from the cross-attention scores of the preceding decoder block to provide accurate and fine-grained positional priors, facilitating locating and refining the target segments progressively.}
  \vspace{-0.5em}
  \label{fig:init}
\end{figure*}

In DETR-like frameworks, providing the queries with meaningful positional priors and encourage each query to concentrate on specific regions is essential to learn representative queries~\cite{wang2021anchor,liu2022dabdetr,zhu2021deformable,yao2021efficient}. In this spirit, masked attention~\cite{cheng2021masked} is proposed, which restricts each query to only attend to a foreground region predicted by the previous decoder block with binary masks. Although promising, the positional priors in masked attention may be inaccurate and deteriorate performance for two reasons. First, each query comprises a content query that contains semantic information and a positional query that provides positional information for the likely locations of the target segments. However, masked attention still relies on positional queries that are randomly initialized learnable parameters~\cite{carion2020end,transformer} (Figure~\ref{fig:init}~(a)), which tend to encode the average statistics across the dataset and cannot reflect the segments with large location variances. Second, since each query only attends to the predicted foreground regions, inaccurate predictions lead to error accumulation across the decoder blocks, especially during an early training stage.

To this end, recent detectors propose to dynamically encode the anchor points into the positional queries to guide queries concentrating around the anchor positions~\cite{meng2021conditional,liu2022dabdetr,wang2021anchor} (Figure~\ref{fig:init}~(b)).
The anchor-based query design mitigates the mentioned issues as the positional queries are dynamically generated for each target object, thus providing more accurate positional priors. In addition, it avoids restricting the queries to only attend to the foreground regions with binary masks to mitigate the error accumulation issue.
However, the anchor-based queries cannot describe the fine-grained positional priors for semantic segmentation, which has details, edges, and boundaries~\cite{chen2017deeplab,chen2018encoder}.

Motivated by the observations that attention scores imply the salient regions for token pruning~\cite{liang2022evit,kong2022spvit}, self-supervised learning~\cite{caron2021emerging}, and semantic segmentation~\cite{zheng2021rethinking,shi2022transformer}, in this paper, we propose a simple yet effective query design for semantic segmentation, dubbed Dynamic Focus-aware Positional Queries (DFPQ), which dynamically generates the positional queries conditioned on the cross-attention scores of the preceding decoder block and the positional encodings for the corresponding image features, simultaneously (Figure~\ref{fig:init}~(c)). In this way, our DFPQ preserves the localization information of the target segments, thereby providing accurate and fine-grained positional priors and facilitating progressively locating and refining the target segments. When implementing the positional encodings with more powerful ones like~\cite{chu2021conditional}, our DFPQ is further empowered with higher capacity to encode the neighbourhood information for the target segments. Compared to the anchor-based positional queries~\cite{liu2022dabdetr,wang2021anchor}, our DFPQ can cover fine-grained locations for the segmentation details, edges, and boundaries which include rich segmentation cues.

In addition, we propose an efficient method named High-Resolution Cross-Attention (HRCA) to mine details for segmenting small regions from the high-resolution feature maps ($1/4\times1/4$ of the original image size). Considering performing cross-attention on high-resolution feature maps requires a formidable amount of memory footprints and computational complexity, \eg, 11G extra floating-point operations with an input resolution of $512\times 512$, we propose to encode token affinity only on the informative areas of high-resolution feature maps that are indicated important in the low-resolution counterparts. In this way, fine-grained details are learned efficiently with affordable memory and computations.

Our main contributions can be summarized as follows:
\begin{itemize}
    \vspace{-0.4em}
    \item We make the pioneering attempt to present a simple yet effective query formulation for semantic segmentation, which provides accurate and fine-grained positional priors to localize the target segments, and mitigates the error accumulation problem while being lightweight with little extra computation.
    \vspace{-0.4em}
    \item We propose an efficient high-resolution cross-attention layer to enrich the segmentation details, which discards the semantically unimportant regions for any target segments in the high-resolution feature maps with affordable memory footprint and computational cost.
    \vspace{-0.4em}
    \item
    Extensive experiments on ADE20K and Cityscapes datasets demonstrate that simply incorporating our DFPQ and HRCA into Mask2former~\cite{cheng2021masked} achieves significant performance gain and outperforms the SOTA methods. For instance, our FASeg outperforms SOTA methods by 1.1\%, 1.3\%, and 0.9\% single-scale mIoU on the ADE20K~\cite{zhou2017scene} validation set with ResNet-50, Swin-T, and Swin-B backbones, respectively. 
\end{itemize}
\vspace{-0.7em}
\section{Related Work}\label{subsec:related_work}
\vspace{-0.5em}

\noindent\textbf{Semantic segmentation with Transformers.}
The recent segmentors with Transformers~\cite{strudel2021segmenter,zhang2021k,yuan2022volo,li2022mask} have pushed the horizon for semantic segmentation.
In general, these segmentors consist of three modules: a backbone, a neck, and a head. Correspondingly, the recent advances can be roughly split into three orthogonal categories. The first category~\cite{dong2021cswin,yuan2022volo,yang2021focal, xu2021vitae, zhang2023vitaev2, zhang2022vsa} aims at learning more representative features by improving the backbone, mostly by improving the self-attention mechanism in Transformers. For example, focal self-attention~\cite{yang2021focal} combines both fine-grained and coarse-grained features in a backbone self-attention layer. To provide better multi-scale features with neck, the second category~\cite{xie2021segformer,shi2022transformer,jain2021semask,huang2021fapn, shi2022transformer} improves the feature pyramid network (FPN)~\cite{lin2017feature} or pyramid scene parsing (PSP)~\cite{zhao2017pyramid} structure. For instance, SegFormer~\cite{xie2021segformer} simplifies FPN under the Transformer backbone to achieve a better accuracy-efficiency trade-off, and SegDeformer~\cite{shi2022transformer} adds external memory tokens to preserve the global information. The third category implements the head with Transformer and conduct set prediction following the DETR-like end-to-end framework~\cite{carion2020end}. In DETR-like framework, target segments are represented by a set of queries. Considering the importance of providing positional priors for the queries~\cite{wang2021anchor,liu2022dabdetr,meng2021conditional}, masked attention~\cite{cheng2021masked} is proposed to restrict the cross-attention only to the local features. Our work also aims at providing better positional priors. In contrast to~\cite{cheng2021masked}, we follow the recent detectors~\cite{liu2022dabdetr,wang2021anchor} to provide accurate positional priors with dynamic positional queries rather than the learnable parameterized positional queries~\cite{carion2020end}. Differently, our DFPQ provides fine-grained positional priors that can cover the locations for fine segmentation details, edges, and boundaries. Very recent work~\cite{li2022mask} proposes a versatile multi-task head structure to share the mutual information among the segmentation and detection tasks, which however, is not directly comparable to our work.

\noindent\textbf{Positional encodings for Transformers.}
Both self-attention and cross-attention for Transformers are permutation-equivalent. Therefore, Positional Encodings (PE) play an essential role in introducing the order of the sequence. In general, the positional encodings include: absolute PE that is generated with sinusoidal functions~\cite{transformer,shiv2019novel} or being entirely learnable parameters~\cite{gehring2017convolutional,swin}; relative PE that encodes distances between the input tokens~\cite{dai2019transformer,shaw2018self}; and conditional PE, which is dynamically generated, \eg, PEG~\cite{chu2021conditional} generates positional encodings with depth-wise convolution conditioned on the local neighbourhood information. In the same spirit, our DFPQ is also dynamically generated by exploring the idea of conditional encoding~\cite{yang2019condconv,tian2020conditional}, thus delivering higher segmentation accuracy. Differently, our DFPQ is tailored specifically for DETR-style semantic segmentation to learn positional priors for each target segment. In addition, since our DFPQ is conditional on the PE for the image features, implementing it with the more powerful PEs~\cite{gehring2017convolutional,chu2021conditional} can further boost the representational capability of our DFPQ. We investigate the effect of different PE strategies in Section~\ref{subsec:abl}.

When solely pre-training Transformer backbones, the positional encodings are generally seen as a part of the feature embeddings and directly be combined with patch features after patchifying the image~\cite{dosovitskiy2021an,touvron2021training}. Differently, in the cross-attention layers of DETR-like frameworks, both the image features and the object queries require additional positional information to provide positional priors for aggregating the query-specific context, which we refer the readers to Section~\ref{subsec:revisit} for details. Recent detectors~\cite{wang2021anchor,liu2022dabdetr} encode anchor positions into positional queries. In contrast, we design a novel positional query formulation for semantic segmentation to reflect regions of interest instead of anchor points to preserve fine segmentation details.
\vspace{-0.5em}
\section{Method} 
\label{subsec:dynamic_queries}

\subsection{Preliminary: Cross-attention in DETR}
\label{subsec:revisit} 
Before introducing our DFPQ, we first revisit the cross-attention layers in DETR-like frameworks~\cite{carion2020end}. Cross-attention layer is a basic module that updates the object queries by aggregating the image context. Since the cross-attention layer is permutation-invariant, both queries and keys require positional information, which introduces the order and provides positional priors to encourage high attention scores for positionally important regions. Specifically, with $N$, $D$, $H$ and $W$ respectively denoting the number of queries, the hidden dimensions, the height, and the width of the image features, we have the image features $\boldsymbol{K}_c$ and their positional encodings $\boldsymbol{K}_p$ and get keys $\boldsymbol{K}=\boldsymbol{K}_c + \boldsymbol{K}_p$, where $\boldsymbol{K} \in \mathbb{R}^{HW \times D}$. We also have the object queries $\boldsymbol{Q} \in \mathbb{R}^{N \times D}$, where each query consists of a content query $\boldsymbol{Q}_c$ and a positional query $\boldsymbol{Q}_p$. 

Then, the cross-attention operation can be formulated as
\begin{equation}
    \begin{aligned}
        \operatorname{Crs-Attention }(\boldsymbol{Q}, \boldsymbol{K}, \boldsymbol{V})=\operatorname{softmax}\left(\frac{\boldsymbol{Q} \boldsymbol{K}^{T}}{\sqrt{D}}\right) \boldsymbol{V},
    \end{aligned}
\label{eq:crs_attention}
\end{equation}
where $\boldsymbol{V} \in \mathbb{R}^{HW \times D}$ is also the image features in the DETR-like frameworks~\cite{zhang2021k,cheng2021per,carion2020end} and we omit all the linear projections and bias terms for simplicity. From Eq.~(\ref{eq:crs_attention}), we can interpret the cross-attention as aggregating image context based on the dot-product similarity between $\boldsymbol{Q}$ and $\boldsymbol{K}$. Since both the content parts and the positional parts for $\boldsymbol{Q}$ and $\boldsymbol{K}$ contribute to calculating the attention scores, similarities for both parts are considered. Therefore, content similarity contributes to mining the correlation between the object queries and the image features, while positional similarity provides positional priors for each target segment.

\begin{figure*}[t!]
  \centering
  \includegraphics[width=1.0\linewidth]{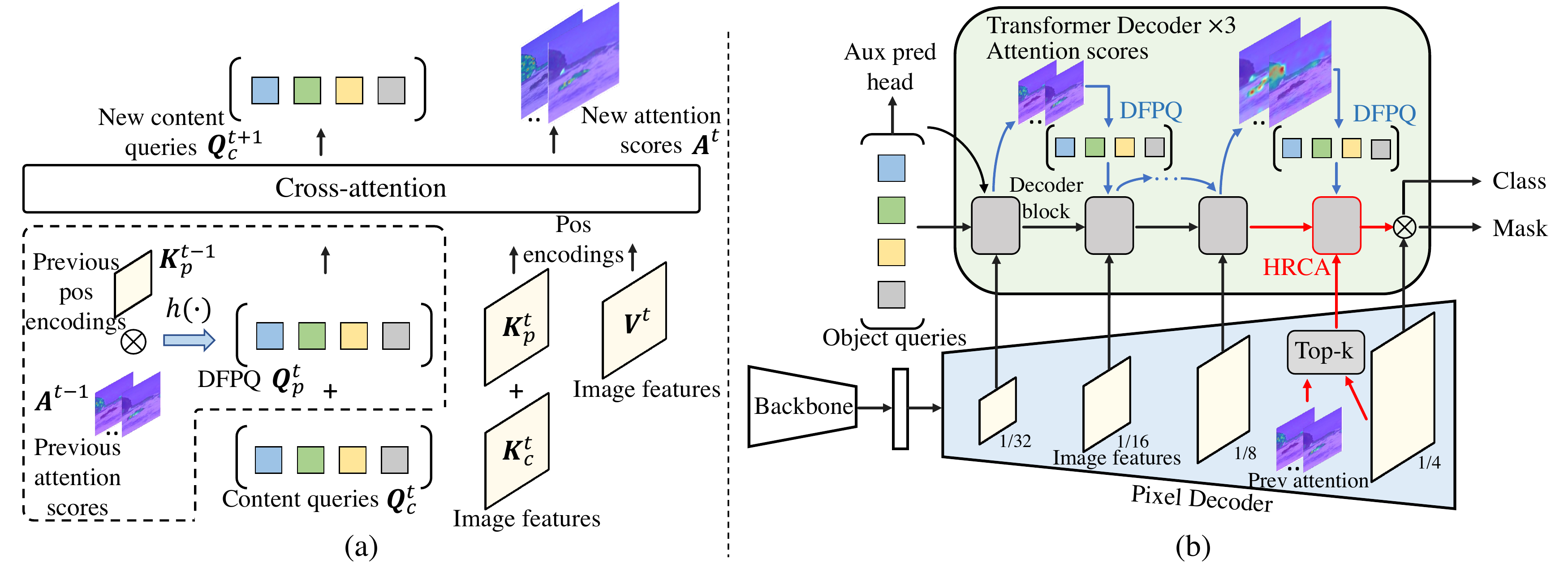}
  \vspace{-2.0em}
  \caption{(a) Cross-attention with our dynamic focus-aware positional queries (DFPQ). ``pos'' is short for positional. We show generating DFPQ in the dashed box, where we multiply the positional encodings for the image features with the cross-attention scores of the preceding decoder block followed by a projection function $h$ to get DFPQ. Here we omit the bias terms for simplicity. (b) The overall framework for our FASeg is built upon Mask2former~\cite{cheng2021masked}, which employs a Backbone to encode images, a Pixel Decoder to fuse the features under different resolutions, and a Transformer Decoder to learn the representation for each target segment. We first apply our DFPQ in each decoder block to provide more accurate positional priors (marked with \textcolor{blue}{blue} arrows). Then, we further propose to incorporate our high-resolution cross-attention (HRCA) layers to model the cross-attention between the queries and the high-resolution feature maps (marked with \textcolor{red}{red} arrows). Here ``Top-k'' selects the top-k pixels indicated by the cross-attention scores of the previous Transformer decoder block.
  }
  \vspace{-1.0em}
  \label{fig:main}
\end{figure*}

\subsection{Dynamic Focus-aware Positional Queries}
\label{subsec:dfpq}

In this work, we aim to develop positional queries that provide effective positional priors under DETR-like frameworks for semantic segmentation. We argue that generating positional queries conditioned on cross-attention scores has three good properties.
First, the cross-attention scores indicate the areas with rich context and may directly reflect the localization information for the target segments~\cite{caron2021emerging,zheng2021rethinking}. Therefore, when stacking several decoder blocks with cross-attention layers in DETR-like frameworks, the localization information in the preceding block is helpful for progressively locating the target segments in the later blocks, especially when the blocks handle features at different scales~\cite{jain2021semask,cheng2021masked}. Second, cross-attention scores are dynamically generated. In contrast to the content-agnostic positional queries as learnable parameters in~\cite{carion2020end}, which tend to encode statistics across the dataset and limit models' generalization capability, cross-attention scores are conditional on each target segment reflecting the specific contextual locations, thereby being more accurate. Finally, the cross-attention scores can cover fine-grained segmentation details, edges, and boundaries instead of encoding only a single center or anchor point alike~\cite{wang2021anchor,liu2022dabdetr}.

Therefore, we propose to generate the positional queries conditional on the cross-attention scores of the preceding decoder block and the positional encodings for the corresponding image features, simultaneously, as shown in Figure~\ref{fig:main}~(a). Specifically, since the positional encodings $\boldsymbol{K}_p$ for the image features preserve the positional information, we form our DFPQ by aggregating $\boldsymbol{K}_p$ as indicated by the cross-attention scores $\boldsymbol{A}$ in the cross-attention layer of the preceding decoder block, which can be formulated as
\begin{equation}
    {\begin{array}{ll}
    \begin{aligned}
        \boldsymbol{Q}^t_p=h(\boldsymbol{A}^{t-1}\boldsymbol{K}_p^{t-1} + \boldsymbol{B}),
    \end{aligned}
    \end{array}}
\label{eq:focus_aware}
\end{equation}
where $t$ is the index of the $t$-th Transformer decoder block, $\boldsymbol{A}^{t-1} \in \mathbb{R}^{N \times HW}$ is the cross-attention scores from the ($t-1$)-th Transformer decoder block, $\boldsymbol{B}\in\mathbb{R}^{N \times D}$ is learnable network parameters, and $h$ is a two-layered MLP with $\mathrm{ReLU}$ non-linearity in between.
Note that the bias term $\boldsymbol{B}$ is the original randomly initialized learnable positional queries, which we employ to stable the training in an early training stage. In this way, we dynamically generate DFPQ to provide positional priors for the target segments. It can also cover the fine-grained segmentation cues that are not restricted by anchor points.

Note that as our DFPQ directly aggregates the positional information, implementing DFPQ with different $\boldsymbol{K}_p$ leads to distinct behaviours. When implementing $\boldsymbol{K}_p$ with the absolute sinusoidal function, the resulting DFPQ reflects an anchor point alike~\cite{liu2022dabdetr,wang2021anchor} instead of the target areas. In this case, we implement $\boldsymbol{K}_p$ with conditional positional encodings~\cite{chu2021conditional} to further encode the neighbourhood information and preserve the implicit positional priors for localizing the target segments. We empirically investigate the effect of different positional encodings in Section~\ref{subsec:abl}.

\subsection{Efficient High-resolution Cross-attention}
\label{subsec:efficient_cross_attention}
As demonstrated by the prior arts~\cite{tian2020conditional,jain2021semask}, high-resolution image features are important for segmenting small regions. However, modelling cross-attention between object queries and high-resolution image features requires an unbearable amount of memory footprints and computational cost. In this case, we propose an efficient High-Resolution Cross-Attention (HRCA) layer to mine details from high-resolution feature maps with affordable memory burden. Specifically, we first select the top-k pixels from the low-resolution image features with the highest cross-attention scores for all object queries. 
We then map these areas to the high-resolution feature map positions in a top-down manner and only perform cross-attention on these positions. Formally, we first get the low-resolution cross-attention scores $\boldsymbol{A}_l$, and then derive its high-resolution counterpart $\boldsymbol{A}_h=f(\boldsymbol{A}_l)$ with bilinear upsampling operation $f(\cdot)$. We next include the top-k pixels in $\boldsymbol{A}_h$ with the highest scores into set $\Omega$, and the efficient HRCA can be formulated as
\begin{equation}
    {\begin{array}{ll}
    \begin{aligned}
        \operatorname{HRCA}(\boldsymbol{Q}, \boldsymbol{K}, \boldsymbol{V}, \Omega)=\operatorname{softmax}\left(\frac{\boldsymbol{Q} \boldsymbol{K}'^{T}}{\sqrt{D}}\right) \boldsymbol{V}',
    \end{aligned}
    \end{array}}
\label{eq:FGCA}
\end{equation}
where $\boldsymbol{K}'=g\left(\boldsymbol{K}, \Omega\right)$, $\boldsymbol{V}'=g(\boldsymbol{V}, \Omega)$ and $g$ is the indexing operation. In this way, we only perform cross-attention on the informative areas for high-resolution feature maps, thereby saving considerable resource consumption.

Our HRCA is closely related to the previous sparse attention methods~\cite{Beltagy2020Longformer,zaheer2020big,tang2022quadtree, wang2022kvt} that only attend to a part of the entire sequence. Differently, our HRCA is specialized to the DETR-like frameworks, which determines the informative pixels based on their contribution to the target segments instead of the other pixels. One similar work to our HRCA is the RCDA module~\cite{wang2021anchor}, which is a representative sparse cross-attention method that decouples cross-attention into a row-wise and column-wise attention to reduce the memory and computation cost. We include the comparison between our HRCA and RCDA~\cite{wang2021anchor} in Section~\ref{subsec:abl}.  

\subsection{Focus-aware Segmentation Framework}\label{subsec:framework}

We first briefly introduce Mask2former~\cite{cheng2021masked}, which consists of backbone, neck, and head as introduced in Section~\ref{subsec:related_work} with the  neck and head named ``Pixel Decoder'' and ``Transformer Decoder'', respectively. In Mask2former, Pixel Decoder fuses the features at multiple scales following~\cite{lin2017feature,zhu2021deformable}. Transformer Decoder cascades three blocks which model the cross-attention between the object queries and the image features of $1/32\times1/32$, $1/16\times1/16$, and $1/8\times1/8$ of the original image resolution, respectively. The Transformer Decoder is repeated three times. We refer readers to~\cite{cheng2021per,cheng2021masked} for more details.

We develop our FASeg upon the Mask2former~\cite{cheng2021masked} framework by simply incorporating our DFPQ and HRCA. The overview of our FASeg is depicted in Figure~\ref{fig:main} (b). We first provide more accurate and fine-grained positional priors for Mask2former with DFPQ (Section~\ref{subsec:dfpq}). We apply DFPQ in the cross-attention layers for each decoder block to provide good positional priors for aggregating the contextual image features to locate target segments. In this way, we progressively localize the target segments as we go deeper in the decoder blocks. Since there are no cross-attention scores before the first Transformer decoder block, we obtain the DFPQ for the first block by performing average pooling on the predicted foreground mask from the auxiliary prediction head as introduced in~\cite{cheng2021masked}.
Next, we employ HRCA (Section~\ref{subsec:efficient_cross_attention}) to enrich the segmentation details with affordable peak-time memory footprints and computational complexity. We add a
fourth decoder block equipped with HRCA to model cross-attention on the high-resolution feature maps after the cascaded three decoder blocks that are already in Mask2former in a top-down manner. With the two simple modifications, our FASeg achieves solid performance gain over the original Mask2former (See Section~\ref{subsec:main_result}).
\section{Experiments}
\label{subsec:experiment}

\begin{table}[t!]
\centering
\caption{Performance comparisons with the state-of-the-art semantic segmentation methods on ADE20K \texttt{val}~\cite{zhou2017scene} with 150 categories. \#P and \#F indicate the number of parameters (M) and FLOPs (G). 
We report both single-scale (s.s.) and multi-scale (m.s.) inference results.
}
\vspace{-0.5em}
\resizebox{\linewidth}{!}{%
\begin{tabular}{c|c|ccccc}
Method & Backbone & \begin{tabular}[c]{@{}c@{}} mIoU \\  s.s. (\%) \end{tabular}    & \begin{tabular}[c]{@{}c@{}} mIoU \\  m.s. (\%) \end{tabular}  & \#P & \#F \\ \shline
UperNet~\cite{xiao2018unified} & R50 & 42.1 & - & 67 & 238 \\
DeepLab V3+~\cite{chen2018encoder} & R50 & 44.0 & 44.9 & 44 & 177  \\
SenFormer~\cite{bousselham2021efficient} & R50 & 44.7 & 45.2 & 144 & 179 \\
Maskformer~\cite{cheng2021per} & R50  & 44.5 & 46.7 & 41 & 53 \\ 
PFD~\cite{qin2022pyramid} & R50  & 45.6 & 48.3 & 74 & 61 \\ 
Mask2former~\cite{cheng2021masked} & R50 & 47.2 & 49.2 & 44 & 71 \\
FASeg (ours) & R50 & \textbf{48.3} & \textbf{49.3} & 51 & 72 \\
\shline
UperNet~\cite{xiao2018unified} & Swin-T & 44.4 & 46.1 & 60 & 236 \\
SenFormer~\cite{bousselham2021efficient} &  Swin-T & 46.0 & - & 144 & 179 \\
Maskformer~\cite{cheng2021per} & Swin-T & 46.7 & 48.8 & 42 & 55 \\
PFD~\cite{qin2022pyramid} & Swin-T & 48.3 & 49.6 & 74 & 65 \\
Mask2former~\cite{cheng2021masked} & Swin-T & 47.7 & 49.6 & 47 & 74 \\
FASeg (ours)& Swin-T & \textbf{49.6} & \textbf{51.3} & 54 & 75 \\\shline
SenFormer~\cite{bousselham2021efficient} &  Swin-B & 51.8 & - & 204 & 242 \\
Maskformer~\cite{cheng2021per} & Swin-B & 52.7 & 53.9 & 102 & 195 \\
PFD~\cite{qin2022pyramid} & Swin-B & 54.1 & 55.3 & 123 & 206 \\
Mask2former~\cite{cheng2021masked} & Swin-B & 53.9 & 55.1 & 107 & 223 \\
FASeg (ours) & Swin-B & \textbf{55.0} & \textbf{56.0} & 113 & 225 \\
\shline
UperNet~\cite{xiao2018unified} & Swin-L & 52.1 & 53.5& 234 & 647  \\
SenFormer~\cite{bousselham2021efficient} &  Swin-L & 53.1 & 54.2 & 314 & 546 \\
Maskformer~\cite{cheng2021per} & Swin-L & 54.1 & 55.6 & 212 & 375 \\
PFD~\cite{qin2022pyramid} & Swin-L & 56.0 & 57.2 & 242 & 385 \\
Mask2former~\cite{cheng2021masked} & Swin-L & 56.1 & 57.3 & 215 & 403 \\
FASeg (ours) & Swin-L & \textbf{56.3} & \textbf{57.7} & 222 & 405 \\
\end{tabular}%
}
\vspace{-0.5em}
\label{tab:ADE_new}
\end{table}

\begin{table}[t!]
\centering
\caption{Performance comparisons with the state-of-the-art semantic segmentation methods on Cityscapes \texttt{val}~\cite{cordts2016cityscapes}. We report single-scale (s.s.) inference results.  \#P and \#F indicate the number of parameters (M) and FLOPs (G). 
}
\vspace{-0.5em}
\resizebox{\linewidth}{!}{%
\begin{tabular}{c|c|ccc}
Method & Backbone & mIoU s.s. (\%) & \#P & \#F \\ \shline
Maskformer~\cite{cheng2021per} & R50 & 78.5 & 41 & 405 \\
Senformer~\cite{bousselham2021efficient} & R50 & 78.8 & 144 & 1,317 \\
DeepLab V3+~\cite{bousselham2021efficient} & R50 & 79.0 & - & - \\
Mask2former~\cite{cheng2021masked} & R50 & 79.4 & 44 & 526 \\
Maskformer~\cite{cheng2021per} & R101 & 79.1 & 60 & 561 \\
Mask2former~\cite{cheng2021masked} & R101 & 80.1 & 67 & 628 \\
SenFormer~\cite{bousselham2021efficient} & R101 & 80.3 & 162 & 1,473 \\
FASeg (ours) & R50 & \textbf{80.5} & 67  & 533 \\
\end{tabular}%
}
\label{tab:cityscapes}
\vspace{-1em}
\end{table}

\noindent\textbf{Implementation details.} Unless otherwise specified, we adopt the same training settings and implementation details as in Mask2former~\cite{cheng2021masked}. For our efficient HRCA in Section~\ref{subsec:efficient_cross_attention}, we select $|\Omega|=\lfloor HW/32 \rfloor$ from the low-resolution feature maps ($1/32\times 1/32$ of the original image size). By default, we train our models with a batch size of 16 on 8 NVIDIA
V100 GPUs. We adopt ResNet~\cite{he2016deep} and Swin Transformer~\cite{swin} pre-trained backbones. For ResNet~\cite{he2016deep}, we use the ResNet-50 (R50) variant. For Swin Transformer~\cite{swin}, we use the Swin-T, Swin-B, and Swin-L backbones where Swin-B and Swin-L are pre-trained on ImageNet-22k~\cite{deng2009imagenet}. Unless specified, we adopt all training settings the same as the default settings of FASeg with R50~\cite{he2016deep} backbone on ADE20K \texttt{val}~\cite{zhou2017scene} with 150 categories for ablation experiments. We conduct the main experiments and ablation studies with the same seeds as Mask2former to seek fair comparisons.

\noindent\textbf{Datasets.} We conduct our experiments on ADE20K~\cite{zhou2017scene} and Cityscapes~\cite{cordts2016cityscapes}. ADE20K~\cite{zhou2017scene} is one of the most challenging large-scale datasets for semantic segmentation, which covers 150 fine-grained semantic concepts, where the training set and validation set contain 20,210 and 2,000 images, respectively. Cityscapes~\cite{cordts2016cityscapes} is an urban street-view dataset with high-resolution images from 50 cities with 19 semantic classes, which consists of 2,975 images for the training set and 2,725 images for the validation set.

\noindent\textbf{Evaluation metrics. } We use single-scale (s.s.) and multi-scale (m.s.) mean Intersection over Union (mIoU)~\cite{everingham2015pascal} as the evaluation metric. We also compare models in terms of their model size (number of parameters) and computational complexity with Floating-point Operations (FLOPs) to evaluate the efficiency of these models. For ablation studies on HRCA, we also show the training-time GPU memory consumption.
For ADE20K~\cite{zhou2017scene} and Cityscapes~\cite{cordts2016cityscapes}, we calculate FLOPs with fixed $512\times 512$ and $1024\times 2048$ image size, respectively.

\noindent\textbf{Compared methods.} We compare our method with the SOTA semantic segmentation methods, including DeepLab V3+~\cite{chen2018encoder}, UperNet~\cite{xiao2018unified}, Maskformer~\cite{cheng2021per}, SenFormer~\cite{bousselham2021efficient}, PFD~\cite{qin2022pyramid} and Mask2former~\cite{cheng2021masked}. Among them, SenFormer~\cite{bousselham2021efficient}, PFD~\cite{qin2022pyramid} and Mask2former~\cite{cheng2021masked} are the recent Transformer-based segmentors, where PFD learns a hierarchy of latent queries to enrich the multi-scale information and SenFormer ensembles the multi-scale predictions. We refer the readers to Section~\ref{subsec:related_work} for more details.

\vspace{-0.2em}
\subsection{Main Results}
\label{subsec:main_result}
\vspace{-0.2em}

We compare our FASeg with state-of-the-art semantic segmentation methods on ADE20K \texttt{val}~\cite{zhou2017scene} and Cityscapes \texttt{val}~\cite{cordts2016cityscapes}. The results are reported in Tables~\ref{tab:ADE_new} and~\ref{tab:cityscapes}. We observe that on ADE20K \texttt{val} (Table~\ref{tab:ADE_new}), with affordable number of extra parameters and FLOPs, our FASeg consistently outperforms the SOTA methods. Specifically, FASeg achieves 48.3\%, 49.6\%, 55.0\%, and 56.3\% mIoU for single-scale inference, outperforming the SOTA methods by 1.1\%, 1.3\%, 0.9\%, and 0.2\% on R50, Swin-T, Swin-B, and Swin-L backbones, respectively. The solid performance gain demonstrates the superiority of our FASeg framework. Our FASeg has more improvements with the smaller backbones (\eg, R50, Swin-T, and Swin-B). We conjecture that localizing the contextural features with smaller backbones under inferior representational capability is challenging. Nevertheless, our DFPQ provides more accurate positional priors, which ease the localization difficulty and lead to better results. For the comparisons on Cityscapes \texttt{val} in Table~\ref{tab:cityscapes},
we observe that with the R50 backbone, our FASeg outperforms all the SOTA methods under desirable numbers of parameters and FLOPs. Surprisingly, FASeg even outperforms the SOTA methods employing the R101 backbone, which demonstrates the effectiveness of our FASeg. To further investigate the flexibility and potential of our main contribution DFPQ, we show more experiments on instance segmentation in the supplementary material.

\begin{figure}[t!]
  \centering
  \includegraphics[width=1.0\linewidth]{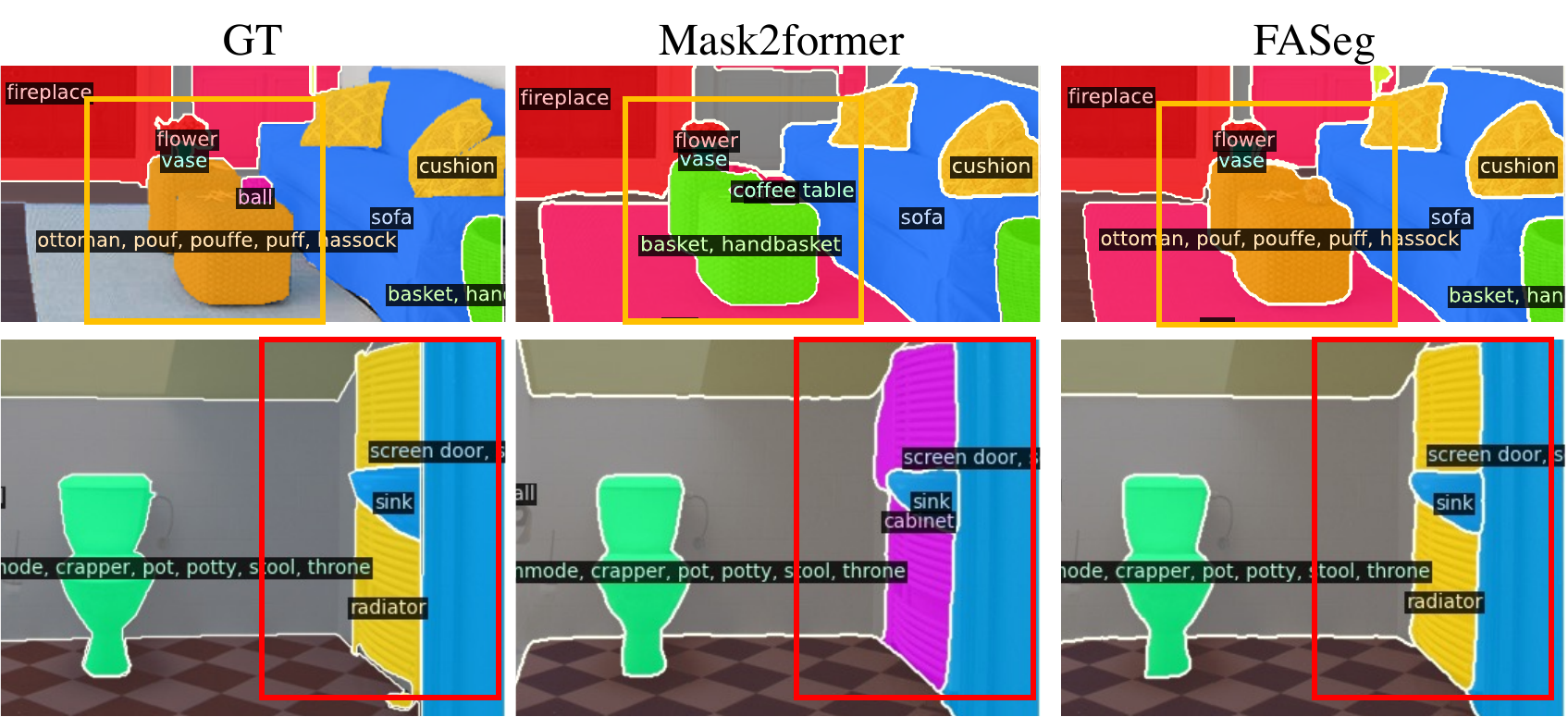}
  \vspace{-2.0em}
  \caption{Qualitative results on the ADE20K \texttt{val}~\cite{zhou2017scene}. Compared to Mask2former~\cite{cheng2021masked}, our FASeg predicts masks with finer details and yields more accurate predictions.}
  \vspace{-1.0em}
  \label{fig:vis}
\end{figure}

We next show some qualitative results in Figure~\ref{fig:vis} and find that our FASeg provides more accurate predictions with finer details. The improved segmentation results again show the superiority of our DFPQ and HRCA. We include more qualitative results in the supplementary material.

\begin{table}[tb!]
\centering
\caption{Effect of the positional encodings $\boldsymbol{K}_p$ for the image features on ADE20K \texttt{val}~\cite{zhou2017scene} with 150 categories.}
\vspace{-0.5em}
\resizebox{0.47\textwidth}{!}{%
\begin{tabular}{c|c|c}
$\boldsymbol{K}_p$ & \begin{tabular}[c]{@{}c@{}} Mask2former~\cite{cheng2021masked} \\ mIoU s.s. (\%) \end{tabular} & \begin{tabular}[c]{@{}c@{}} FASeg \\ mIoU s.s. (\%) \end{tabular} \\ \shline
Sinusoidal~\cite{carion2020end} & 47.2 & 46.9 \\
Learnable absolute~\cite{gehring2017convolutional} &  47.0 & 47.5 \\
Conditional~\cite{chu2021conditional} & 47.3 & 48.3 \\
\end{tabular}%
}
\label{tab:k_p}
\vspace{-0.8em}
\end{table}

\begin{table}[t!]
\centering
\caption{Ablation study for FASeg  on ADE20K \texttt{val}~\cite{zhou2017scene} and Cityscapes \texttt{val}~\cite{cordts2016cityscapes}. \#P and \#F indicate the number of parameters (M) and
FLOPs (G) evaluated on 512$\times$512 images.}
\vspace{-0.5em}
\resizebox{0.47\textwidth}{!}{%
\begin{tabular}{cc|cccc}
DFPQ & HRCA & \begin{tabular}[c]{@{}c@{}} ADE20K \texttt{val} \\ mIoU s.s. (\%) \end{tabular} & \begin{tabular}[c]{@{}c@{}} Cityscapes \texttt{val} \\ mIoU s.s. (\%) \end{tabular}  & \#P & \#F \\ \shline
 &  & 47.2 & 79.4 & 44 & 71  \\
\checkmark &  & 47.7 & 80.0 & 44 & 71 \\
 & \checkmark & 47.6 & 79.8 & 50 & 72 \\
\checkmark & \checkmark & 48.3 & 80.5 & 51 & 72 \\
\end{tabular}%
}
\vspace{-1em}
\label{tab:each_module}
\end{table}

\begin{table}[t!]
\centering
\caption{Performance comparisons between DFPQ and other positional queries variants on ADE20K \texttt{val}~\cite{zhou2017scene} with 150 categories.}
\vspace{-0.5em}
\resizebox{0.45\textwidth}{!}{%
\begin{tabular}{c|c}
Method & mIoU s.s.(\%) \\ \shline
Learnable positional queries & 46.9  \\
Pre-defined grid anchor positional queries & 46.6  \\
Dynamic anchor positional queries &  47.0 \\
Dynamic foreground positional queries & 47.8  \\
DFPQ & 48.3  \\
\end{tabular}%
}
\label{tab:abl_positional}
\vspace{-1em}
\end{table}

\subsection{Ablation Study}\label{subsec:abl}

\noindent\textbf{Effect of $\boldsymbol{K}_p$.} We investigate the effect of the positional encodings $\boldsymbol{K}_p$ for the image features on ADE20k \texttt{val} with R50 backbone. The results are reported in Table~\ref{tab:k_p}. We observe that different $\boldsymbol{K}_p$ have similar performance for Mask2former~\cite{cheng2021masked}. However, more powerful $\boldsymbol{K}_p$ leads to much higher performance for our FASeg. For instance, FASeg with conditional positional encodings~\cite{chu2021conditional} outperforms Mask2former counterpart and FASeg with sinusoidal positional encodings~\cite{carion2020end} by 1.0\% and 1.4\% mIoU, respectively. The reason is that compared to Mask2former, our FASeg additionally aggregates $\boldsymbol{K}_p$ to get DFPQ as explained in Section~\ref{subsec:dfpq}. Therefore, more powerful $\boldsymbol{K}_p$ leads to higher representational capability of DFPQ that boosts the performance. We also find that with sinusoidal positional encodings, FASeg has even lower performance than Mask2former as the DFPQ aggregated from sinusoidal positional encodings reflects a single anchor point which cannot cover the fine-grained segmentation cues.

\noindent\textbf{Effectiveness of DFPQ and HRCA.} We investigate the effectiveness of our DFPQ and HRCA on ADE20k \texttt{val} and Cityscapes \texttt{val} with the ResNet-50 backbone. The results are reported in Table~\ref{tab:each_module}. We observe that both DFPQ and HRCA gain clear margins from the vanilla Mask2former~\cite{cheng2021masked}.
To be specific, integrating DFPQ on Mask2former boosts the performance by 0.5\% and 0.6\% mIoU on ADE20k \texttt{val} and Cityscapes \texttt{val}, respectively, with barely any extra parameter and computational cost. It is indicated that DFPQ is lightweight and contributes largely on the performance gain. Employing HRCA on Mask2former leads to 0.4\% mIoU gain on both datasets, which however, has 6M more parameters and 1G higher FLOPs. The additional parameters and FLOPs are brought by the extra decoder layers handling high-resolution image features. Finally, our FASeg with both DFPQ and HRCA improves 1.1\% mIoU for both ADE20k \texttt{val} and Cityscapes \texttt{val}, demonstrating the superiority of our FASeg.

\begin{figure*}[ht!]
  \centering
  \includegraphics[width=1.0\linewidth]{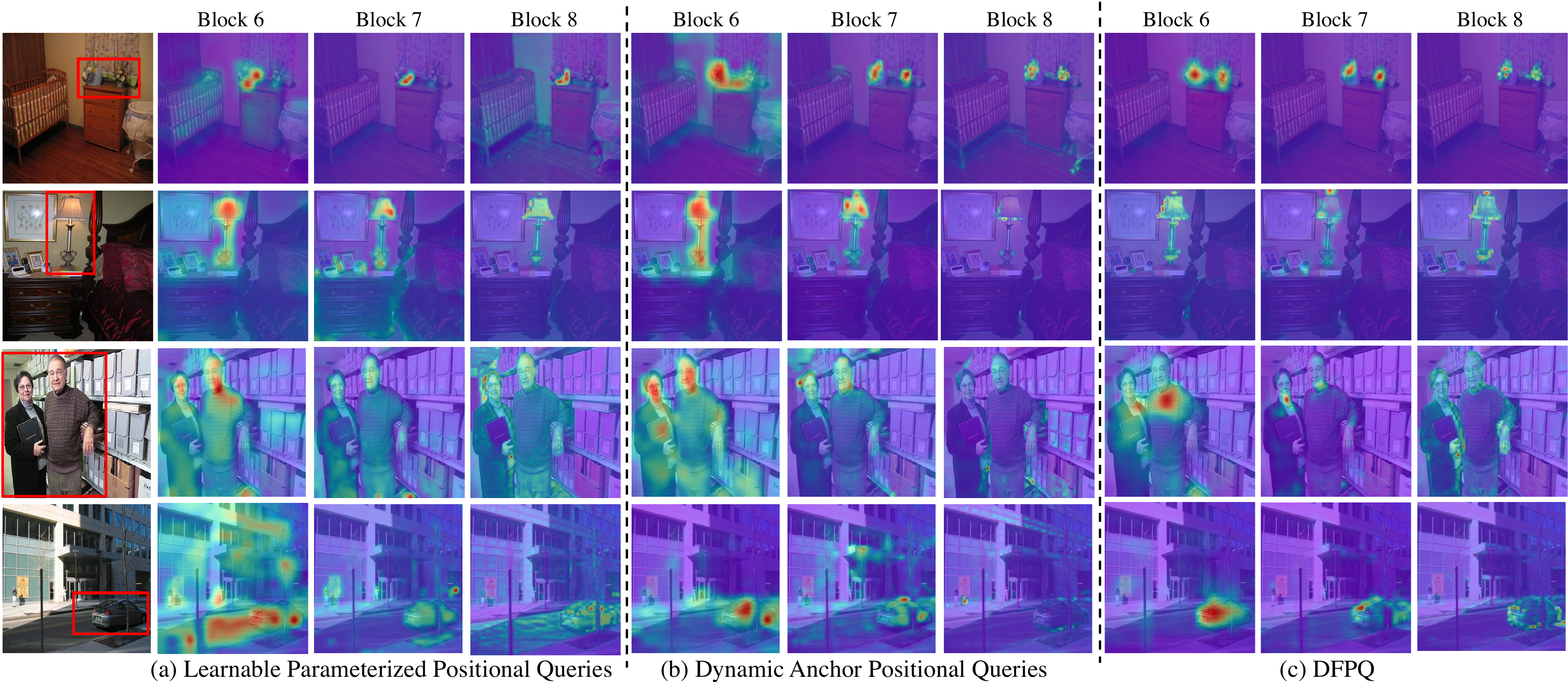}
  \vspace{-2.0em}
  \caption{Visualizations of the cross-attention maps for learnable positional queries (\cite{carion2020end,cheng2021masked}), dynamic anchor positional queries (alike \cite{liu2022dabdetr}) and our DFPQ. We show the visualizations for the normalized cross-attention maps in the last three decoder blocks and indicate the target segments in the red boxes. The cross-attention maps with the learnable positional queries and the dynamic anchor positional queries are often scattered without a clear focus and mix up different segments, while the cross-attention maps with DFPQ are more compact and consistent to reflect the target segments.}
  \vspace{-1em}
  \label{fig:attention}
\end{figure*}

\noindent\textbf{DFPQ vs. other positional queries.} We investigate the effectiveness of our DFPQ and compare it with other learnable query variants on ADE20K \texttt{val}~\cite{zhou2017scene}. The results are presented in Table~\ref{tab:abl_positional}. Here we adopt all the other settings the same as our FASeg with the R50 backbone and only differ the positional queries for all the competitors. Specifically, we compare with four settings: 1) learnable parameterized positional queries that are randomly initialized~\cite{carion2020end}; 2) positional queries as the pre-defined grid anchor points akin to~\cite{wang2021anchor}; 3) positional queries dynamically generated from the center of the foreground masks predicted by the previous layer similar to~\cite{liu2022dabdetr}; 4) positional queries dynamically generated from the entire predicted foreground masks. We find that our DFPQ outperforms all the competitors by large margins. For example, our DFPQ achieves 1.7\% and 1.3\% higher mIoU than the pre-defined grid and dynamic anchor positional queries, respectively. It is suggested that our DFPQ better suits semantic segmentation than the other positional query variants. We also visualize cross-attention maps among the different positional queries in Figure~\ref{fig:attention}. We observe that our DFPQ (Figure~\ref{fig:attention}~(c)) helps generate more compact and consistent cross-attention maps focusing on the target segments than the learnable parameterized positional queries (Figure~\ref{fig:attention}~(a)) and dynamic anchor positional queries (Figure~\ref{fig:attention}~(b)).

\begin{table}[t!]
\centering
\caption{Performance comparisons between our HRCA and other efficient cross-attention methods on ADE20K \texttt{val}~\cite{zhou2017scene} with 150 categories. \#F denotes the number of FLOPs (G). The training memory footprint (M) and FLOPs are measured under 512$\times$512 image resolutions with a batch size of 4 on a single GPU.}
\vspace{-0.5em}
\resizebox{0.47\textwidth}{!}{%
\begin{tabular}{c|c|c|c}
Method& mIoU s.s. (\%) & Training Memory (M)  & \#F \\ \shline
Vanilla & 47.3 & 7,451 & 83 \\
Random $\Omega$ & 46.7 & 6,343 & 72 \\
RCDA &  47.5 & 6,082 & 72  \\
HRCA & 48.3 & 6,343 & 72 \\
\end{tabular}%
}
\label{tab:abl_hrca}
\vspace{-1em}
\end{table}

\begin{table}[t!]
\centering
\caption{Effect of $|\Omega|$ in our efficient HRCA on ADE20K \texttt{val}~\cite{zhou2017scene} with 150 categories. \#F indicates the number of FLOPs (G).}
\vspace{-0.5em}
\resizebox{0.47\textwidth}{!}{%
\begin{tabular}{c|ccc}
$|\Omega|$ & mIoU s.s. (\%) & Training memory (M) & \#F \\ \shline
$HW$ & 47.3 & 7,451 & 83 \\
$\lfloor HW / 16 \rfloor$ &  47.7 & 6,381 & 72  \\
$\lfloor HW / 32 \rfloor$ & 48.3 & 6,343 & 72 \\
$\lfloor HW / 64 \rfloor$ & 48.0 & 6,317 & 71 \\
\end{tabular}%
}
\label{tab:abl_cross}
  \vspace{-1em}
\end{table}

\noindent\textbf{HRCA vs. other efficient cross-attention methods.} We investigate the effectiveness of our HRCA and compare it with other cross-attention methods on ADE20k \texttt{val}. The results are reported in Table~\ref{tab:abl_hrca}. For a fair comparison, we only replace HRCA for the other efficient cross-attention methods on our FASeg with the R50 backbone. We compare with three baselines: 1) the vanilla cross-attention that models the entire high-resolution features; 2) our HRCA with randomly sampled top-k pixels to form set $\Omega$ that $|\Omega|$ is the same as HRCA; 3) RCDA~\cite{wang2021anchor} that the cross-attention is decoupled to a row-wise and column-wise attention as introduced in Section~\ref{subsec:efficient_cross_attention}. We empirically find that compared with the vanilla cross-attention, our HRCA achieves 1.0\% mIoU gain while exhibiting 1,108M lower training-time GPU memory and 11G lower FLOPs. Our HRCA also outperforms the two efficient cross-attention methods by large margins. For example, HRCA achieves 0.8\% higher mIoU than RCDA with marginally increased training-time memory. The results demonstrate the superiority of our HRCA for efficiently identifying and utilizing contextual tokens in high-resolution features.

\noindent\textbf{Effect of $|\Omega|$ in HRCA.} We then investigate how $|\Omega|$ affects the performance, memory consumption and computational complexity on FASeg with R50 backbone on ADE20k \texttt{val}. The results are reported in Table~\ref{tab:abl_cross}. $|\Omega|$ determines the number of contextual tokens used in attention as introduced in Section~\ref{subsec:efficient_cross_attention}. Here we measure the memory consumption by the training-time memory with a batch size of 4 on a single GPU. In the vanilla cross-attention layers, cross-attention attends to the entire feature maps from the encoder, in which case $|\Omega|=HW$. We observe that our HRCA outperforms the vanilla cross-attention by a significant margin.  We conjecture that the sparse property~\cite{frankle2018the,evci2020rigging} has reduced the redundancy in high-resolution feature maps in our HRCA and leads to higher performance and efficiency. Since our HRCA achieves the highest performance when $|\Omega|=\lfloor HW / 32 \rfloor$, we set $|\Omega|=\lfloor HW / 32 \rfloor$ by default for all the other experiments.

\begin{table}[tb!]
\centering
\caption{Effect of applying HRCA to other high-resolution feature scales for FASeg with Swin-B Backbone on ADE20K \texttt{val}~\cite{zhou2017scene} with 150 categories.}
\vspace{-0.5em}
\resizebox{\linewidth}{!}{%
\begin{tabular}{cc|cc}
$1/4\times1/4$ & $1/8\times1/8$ & mIoU s.s. (\%) & Training Memory (M) \\ \shline
\checkmark & & 55.0 & 20,418 \\
\checkmark & \checkmark & 54.9 & 19,898 \\
  & \checkmark & 54.8 & 17,817 \\
\end{tabular}%
}
\label{tab:abl_more_hrca}
\vspace{-2em}
\end{table}

\noindent\textbf{Effect of applying HRCA to other high-resolution feature scales.} By default, HRCA is applied only to the high-resolution $1/4\times1/4$ feature scale. We explore applying HRCA to $1/8\times1/8$ and both $1/4\times1/4$ and $1/8\times1/8$ feature scales for FASeg with Swin-B backbone on ADE20k \texttt{val}. We measure the training-time memory consumption with a batch size of 4 on
a single GPU and report the results in Table~\ref{tab:abl_more_hrca}. We find that the performance only fluctuates within 0.2\% mIoU. In particular, modeling the cross-attention only on the $1/8\times1/8$ feature scale with HRCA saves more than 2,000M training-time memory, suggesting the potential for extending HRCA to more high-resolution features to alleviate the memory burden.


\vspace{-0.6em}
\section{Conclusion}
\vspace{-0.6em}
In this paper, we have explored providing positional priors with positional queries for the DETR-style semantic segmentation. Specifically, we have proposed to dynamically generate the positional queries conditioned on the cross-attention scores of the preceding decoder block and the positional encodings for the corresponding image features, simultaneously. We have found that our novel query design delivers more accurate and fine-grained positional priors facilitating localizing the target segments progressively. To mitigate the training-time memory cost when modeling cross-attention on high-resolution feature maps, we have presented an efficient approach to only aggregate the contextual tokens from the high-resolution feature maps, which is shown to learn low-level details with affordable memory and computations. Finally, we have conducted extensive experiments to demonstrate the effectiveness of our proposed framework on the semantic segmentation task and its potential to be extended to other segmentation tasks.

\noindent\textbf{Limitations and societal impact.} 
Although our HRCA enriches the segmentation details with affordable memory and computations, it still requires more parameters. To this end, we will explore slimming~\cite{chen2021chasing,he2021pruning} or reusing~\cite{shen2022sliced} these blocks to save parameters. Another potential future direction is to explore the explainability of the positional priors generated by our DFPQ. Our technical innovations do not appear to have any negative societal impacts. However, the trained model may deliver unstable or biased predictions with training data that is not reviewed properly.

\noindent\textbf{Acknowledgement.} Dr Jing Zhang was supported by Australian Research Council Projects in part by FL170100117 and IH180100002.

\newpage
{\small
\bibliographystyle{ieee_fullname}
\bibliography{egbib}

\begin{thebibliography}{10}\itemsep=-1pt

\bibitem{Beltagy2020Longformer}
Iz Beltagy, Matthew~E. Peters, and Arman Cohan.
\newblock Longformer: The long-document transformer.
\newblock {\em arXiv:2004.05150}, 2020.

\bibitem{bousselham2021efficient}
Walid Bousselham, Guillaume Thibault, Lucas Pagano, Archana Machireddy, Joe
  Gray, Young~Hwan Chang, and Xubo Song.
\newblock Efficient self-ensemble framework for semantic segmentation.
\newblock {\em arXiv preprint arXiv:2111.13280}, 2021.

\bibitem{carion2020end}
Nicolas Carion, Francisco Massa, Gabriel Synnaeve, Nicolas Usunier, Alexander
  Kirillov, and Sergey Zagoruyko.
\newblock End-to-end object detection with transformers.
\newblock In {\em ECCV}, pages 213--229, 2020.

\bibitem{caron2021emerging}
Mathilde Caron, Hugo Touvron, Ishan Misra, Herv{\'e} J{\'e}gou, Julien Mairal,
  Piotr Bojanowski, and Armand Joulin.
\newblock Emerging properties in self-supervised vision transformers.
\newblock In {\em ICCV}, pages 9650--9660, 2021.

\bibitem{chen2019hybrid}
Kai Chen, Jiangmiao Pang, Jiaqi Wang, Yu Xiong, Xiaoxiao Li, Shuyang Sun,
  Wansen Feng, Ziwei Liu, Jianping Shi, Wanli Ouyang, et~al.
\newblock Hybrid task cascade for instance segmentation.
\newblock In {\em CVPR}, pages 4974--4983, 2019.

\bibitem{chen2017deeplab}
Liang-Chieh Chen, George Papandreou, Iasonas Kokkinos, Kevin Murphy, and Alan~L
  Yuille.
\newblock Deeplab: Semantic image segmentation with deep convolutional nets,
  atrous convolution, and fully connected crfs.
\newblock {\em TPAMI}, 40(4):834--848, 2017.

\bibitem{chen2018encoder}
Liang-Chieh Chen, Yukun Zhu, George Papandreou, Florian Schroff, and Hartwig
  Adam.
\newblock Encoder-decoder with atrous separable convolution for semantic image
  segmentation.
\newblock In {\em ECCV}, pages 801--818, 2018.

\bibitem{chen2021chasing}
Tianlong Chen, Yu Cheng, Zhe Gan, Lu Yuan, Lei Zhang, and Zhangyang Wang.
\newblock Chasing sparsity in vision transformers: An end-to-end exploration.
\newblock {\em NeurIPS}, 34:19974--19988, 2021.

\bibitem{cheng2021masked}
Bowen Cheng, Ishan Misra, Alexander~G Schwing, Alexander Kirillov, and Rohit
  Girdhar.
\newblock Masked-attention mask transformer for universal image segmentation.
\newblock In {\em CVPR}, 2022.

\bibitem{cheng2021per}
Bowen Cheng, Alex Schwing, and Alexander Kirillov.
\newblock Per-pixel classification is not all you need for semantic
  segmentation.
\newblock In {\em NeurIPS}, volume~34, 2021.

\bibitem{chu2021conditional}
Xiangxiang Chu, Zhi Tian, Bo Zhang, Xinlong Wang, Xiaolin Wei, Huaxia Xia, and
  Chunhua Shen.
\newblock Conditional positional encodings for vision transformers.
\newblock In {\em ICLR}, 2023.

\bibitem{cordts2016cityscapes}
Marius Cordts, Mohamed Omran, Sebastian Ramos, Timo Rehfeld, Markus Enzweiler,
  Rodrigo Benenson, Uwe Franke, Stefan Roth, and Bernt Schiele.
\newblock The cityscapes dataset for semantic urban scene understanding.
\newblock In {\em CVPR}, pages 3213--3223, 2016.

\bibitem{dai2019transformer}
Zihang Dai, Zhilin Yang, Yiming Yang, Jaime Carbonell, Quoc~V Le, and Ruslan
  Salakhutdinov.
\newblock Transformer-xl: Attentive language models beyond a fixed-length
  context.
\newblock {\em arXiv preprint arXiv:1901.02860}, 2019.

\bibitem{deng2009imagenet}
Jia Deng, Wei Dong, Richard Socher, Li-Jia Li, Kai Li, and Li Fei-Fei.
\newblock Imagenet: A large-scale hierarchical image database.
\newblock In {\em CVPR}, pages 248--255, 2009.

\bibitem{dong2021cswin}
Xiaoyi Dong, Jianmin Bao, Dongdong Chen, Weiming Zhang, Nenghai Yu, Lu Yuan,
  Dong Chen, and Baining Guo.
\newblock Cswin transformer: A general vision transformer backbone with
  cross-shaped windows.
\newblock In {\em CVPR}, 2022.

\bibitem{dosovitskiy2021an}
Alexey Dosovitskiy, Lucas Beyer, Alexander Kolesnikov, Dirk Weissenborn,
  Xiaohua Zhai, Thomas Unterthiner, Mostafa Dehghani, Matthias Minderer, Georg
  Heigold, Sylvain Gelly, Jakob Uszkoreit, and Neil Houlsby.
\newblock An image is worth 16x16 words: Transformers for image recognition at
  scale.
\newblock In {\em ICLR}, 2021.

\bibitem{evci2020rigging}
Utku Evci, Trevor Gale, Jacob Menick, Pablo~Samuel Castro, and Erich Elsen.
\newblock Rigging the lottery: Making all tickets winners.
\newblock In {\em ICML}, pages 2943--2952, 2020.

\bibitem{everingham2015pascal}
Mark Everingham, SM Eslami, Luc Van~Gool, Christopher~KI Williams, John Winn,
  and Andrew Zisserman.
\newblock The pascal visual object classes challenge: A retrospective.
\newblock {\em IJCV}, 111(1):98--136, 2015.

\bibitem{frankle2018the}
Jonathan Frankle and Michael Carbin.
\newblock The lottery ticket hypothesis: Finding sparse, trainable neural
  networks.
\newblock In {\em ICLR}, 2019.

\bibitem{gehring2017convolutional}
Jonas Gehring, Michael Auli, David Grangier, Denis Yarats, and Yann~N Dauphin.
\newblock Convolutional sequence to sequence learning.
\newblock In {\em ICML}, pages 1243--1252, 2017.

\bibitem{he2021pruning}
Haoyu He, Jing Liu, Zizheng Pan, Jianfei Cai, Jing Zhang, Dacheng Tao, and
  Bohan Zhuang.
\newblock Pruning self-attentions into convolutional layers in single path.
\newblock {\em arXiv preprint arXiv:2111.11802}, 2021.

\bibitem{he2017mask}
Kaiming He, Georgia Gkioxari, Piotr Doll{\'a}r, and Ross Girshick.
\newblock Mask r-cnn.
\newblock In {\em ICCV}, pages 2961--2969, 2017.

\bibitem{he2016deep}
Kaiming He, Xiangyu Zhang, Shaoqing Ren, and Jian Sun.
\newblock Deep residual learning for image recognition.
\newblock In {\em CVPR}, pages 770--778, 2016.

\bibitem{huang2021fapn}
Shihua Huang, Zhichao Lu, Ran Cheng, and Cheng He.
\newblock Fapn: Feature-aligned pyramid network for dense image prediction.
\newblock In {\em ICCV}, pages 864--873, 2021.

\bibitem{jain2021semask}
Jitesh Jain, Anukriti Singh, Nikita Orlov, Zilong Huang, Jiachen Li, Steven
  Walton, and Humphrey Shi.
\newblock Semask: Semantically masked transformers for semantic segmentation.
\newblock {\em arXiv preprint arXiv:2112.12782}, 2021.

\bibitem{kong2022spvit}
Zhenglun Kong, Peiyan Dong, Xiaolong Ma, Xin Meng, Wei Niu, Mengshu Sun, Xuan
  Shen, Geng Yuan, Bin Ren, Hao Tang, et~al.
\newblock Spvit: Enabling faster vision transformers via latency-aware soft
  token pruning.
\newblock In {\em ECCV}, pages 620--640, 2022.

\bibitem{li2022mask}
Feng Li, Hao Zhang, Huaizhe xu, Shilong Liu, Lei Zhang, Lionel~M. Ni, and
  Heung-Yeung Shum.
\newblock Mask dino: Towards a unified transformer-based framework for object
  detection and segmentation.
\newblock {\em arXiv preprint arXiv:2206.02777}, 2022.

\bibitem{liang2022evit}
Youwei Liang, Chongjian Ge, Zhan Tong, Yibing Song, Jue Wang, and Pengtao Xie.
\newblock Not all patches are what you need: Expediting vision transformers via
  token reorganizations.
\newblock In {\em ICLR}, 2022.

\bibitem{lin2017feature}
Tsung-Yi Lin, Piotr Doll{\'a}r, Ross Girshick, Kaiming He, Bharath Hariharan,
  and Serge Belongie.
\newblock Feature pyramid networks for object detection.
\newblock In {\em CVPR}, pages 2117--2125, 2017.

\bibitem{lin2014microsoft}
Tsung-Yi Lin, Michael Maire, Serge Belongie, James Hays, Pietro Perona, Deva
  Ramanan, Piotr Doll{\'a}r, and C~Lawrence Zitnick.
\newblock Microsoft coco: Common objects in context.
\newblock In {\em ECCV}, pages 740--755, 2014.

\bibitem{liu2022dabdetr}
Shilong Liu, Feng Li, Hao Zhang, Xiao Yang, Xianbiao Qi, Hang Su, Jun Zhu, and
  Lei Zhang.
\newblock {DAB}-{DETR}: Dynamic anchor boxes are better queries for {DETR}.
\newblock In {\em ICLR}, 2022.

\bibitem{swin}
Ze Liu, Yutong Lin, Yue Cao, Han Hu, Yixuan Wei, Zheng Zhang, Stephen Lin, and
  Baining Guo.
\newblock Swin transformer: Hierarchical vision transformer using shifted
  windows.
\newblock In {\em ICCV}, 2021.

\bibitem{meng2021conditional}
Depu Meng, Xiaokang Chen, Zejia Fan, Gang Zeng, Houqiang Li, Yuhui Yuan, Lei
  Sun, and Jingdong Wang.
\newblock Conditional detr for fast training convergence.
\newblock In {\em ICCV}, pages 3651--3660, 2021.

\bibitem{qin2022pyramid}
Zipeng Qin, Jianbo Liu, Xiaolin Zhang, Maoqing Tian, Aojun Zhou, Shuai Yi, and
  Hongsheng Li.
\newblock Pyramid fusion transformer for semantic segmentation.
\newblock {\em arXiv preprint arXiv:2201.04019}, 2022.

\bibitem{shaw2018self}
Peter Shaw, Jakob Uszkoreit, and Ashish Vaswani.
\newblock Self-attention with relative position representations.
\newblock In {\em NAACL}, pages 464--468, 2018.

\bibitem{shen2022sliced}
Zhiqiang Shen, Zechun Liu, and Eric Xing.
\newblock Sliced recursive transformer.
\newblock In {\em ECCV}, pages 727--744, 2022.

\bibitem{shi2022transformer}
Hengcan Shi, Munawar Hayat, and Jianfei Cai.
\newblock Transformer scale gate for semantic segmentation.
\newblock In {\em CVPR}, 2023.

\bibitem{shiv2019novel}
Vighnesh Shiv and Chris Quirk.
\newblock Novel positional encodings to enable tree-based transformers.
\newblock {\em NeurIPS}, 32, 2019.

\bibitem{strudel2021segmenter}
Robin Strudel, Ricardo Garcia, Ivan Laptev, and Cordelia Schmid.
\newblock Segmenter: Transformer for semantic segmentation.
\newblock In {\em ICCV}, pages 7262--7272, 2021.

\bibitem{tang2022quadtree}
Shitao Tang, Jiahui Zhang, Siyu Zhu, and Ping Tan.
\newblock Quadtree attention for vision transformers.
\newblock In {\em ICLR}, 2022.

\bibitem{tian2020conditional}
Zhi Tian, Chunhua Shen, and Hao Chen.
\newblock Conditional convolutions for instance segmentation.
\newblock In {\em ECCV}, pages 282--298, 2020.

\bibitem{touvron2021training}
Hugo Touvron, Matthieu Cord, Matthijs Douze, Francisco Massa, Alexandre
  Sablayrolles, and Herv{\'e} J{\'e}gou.
\newblock Training data-efficient image transformers \& distillation through
  attention.
\newblock In {\em ICML}, pages 10347--10357, 2021.

\bibitem{transformer}
Ashish Vaswani, Noam Shazeer, Niki Parmar, Jakob Uszkoreit, Llion Jones,
  Aidan~N. Gomez, Lukasz Kaiser, and Illia Polosukhin.
\newblock Attention is all you need.
\newblock In {\em NeurIPS}, pages 5998--6008, 2017.

\bibitem{wang2022kvt}
Pichao Wang, Xue Wang, Fan Wang, Ming Lin, Shuning Chang, Hao Li, and Rong Jin.
\newblock Kvt: k-nn attention for boosting vision transformers.
\newblock In {\em ECCV}, pages 285--302, 2022.

\bibitem{wang2022towards}
Wen Wang, Jing Zhang, Yang Cao, Yongliang Shen, and Dacheng Tao.
\newblock Towards data-efficient detection transformers.
\newblock In {\em ECCV}, pages 88--105, 2022.

\bibitem{wang2020solov2}
Xinlong Wang, Rufeng Zhang, Tao Kong, Lei Li, and Chunhua Shen.
\newblock Solov2: Dynamic and fast instance segmentation.
\newblock {\em NeurIPS}, 33:17721--17732, 2020.

\bibitem{wang2021anchor}
Yingming Wang, Xiangyu Zhang, Tong Yang, and Jian Sun.
\newblock Anchor detr: Query design for transformer-based detector.
\newblock In {\em AAAI}, 2022.

\bibitem{xiao2018unified}
Tete Xiao, Yingcheng Liu, Bolei Zhou, Yuning Jiang, and Jian Sun.
\newblock Unified perceptual parsing for scene understanding.
\newblock In {\em ECCV}, pages 418--434, 2018.

\bibitem{xie2021segformer}
Enze Xie, Wenhai Wang, Zhiding Yu, Anima Anandkumar, Jose~M Alvarez, and Ping
  Luo.
\newblock Segformer: Simple and efficient design for semantic segmentation with
  transformers.
\newblock In {\em NeurIPS}, volume~34, 2021.

\bibitem{xu2021vitae}
Yufei Xu, Qiming Zhang, Jing Zhang, and Dacheng Tao.
\newblock Vitae: Vision transformer advanced by exploring intrinsic inductive
  bias.
\newblock {\em NeurIPS}, 34:28522--28535, 2021.

\bibitem{yang2019condconv}
Brandon Yang, Gabriel Bender, Quoc~V Le, and Jiquan Ngiam.
\newblock Condconv: Conditionally parameterized convolutions for efficient
  inference.
\newblock In {\em NeurIPS}, volume~32, 2019.

\bibitem{yang2021focal}
Jianwei Yang, Chunyuan Li, Pengchuan Zhang, Xiyang Dai, Bin Xiao, Lu Yuan, and
  Jianfeng Gao.
\newblock Focal self-attention for local-global interactions in vision
  transformers.
\newblock In {\em NeurIPS}, 2021.

\bibitem{yao2021efficient}
Zhuyu Yao, Jiangbo Ai, Boxun Li, and Chi Zhang.
\newblock Efficient detr: improving end-to-end object detector with dense
  prior.
\newblock {\em arXiv preprint arXiv:2104.01318}, 2021.

\bibitem{yuan2022volo}
Li Yuan, Qibin Hou, Zihang Jiang, Jiashi Feng, and Shuicheng Yan.
\newblock Volo: Vision outlooker for visual recognition.
\newblock {\em TPAMI}, 2022.

\bibitem{zaheer2020big}
Manzil Zaheer, Guru Guruganesh, Kumar~Avinava Dubey, Joshua Ainslie, Chris
  Alberti, Santiago Ontanon, Philip Pham, Anirudh Ravula, Qifan Wang, Li Yang,
  et~al.
\newblock Big bird: Transformers for longer sequences.
\newblock In {\em NeurIPS}, volume~33, pages 17283--17297, 2020.

\bibitem{zhang2022vsa}
Qiming Zhang, Yufei Xu, Jing Zhang, and Dacheng Tao.
\newblock Vsa: learning varied-size window attention in vision transformers.
\newblock In {\em ECCV}, pages 466--483, 2022.

\bibitem{zhang2023vitaev2}
Qiming Zhang, Yufei Xu, Jing Zhang, and Dacheng Tao.
\newblock Vitaev2: Vision transformer advanced by exploring inductive bias for
  image recognition and beyond.
\newblock {\em IJCV}, pages 1--22, 2023.

\bibitem{zhang2021k}
Wenwei Zhang, Jiangmiao Pang, Kai Chen, and Chen~Change Loy.
\newblock K-net: Towards unified image segmentation.
\newblock In {\em NeurIPS}, volume~34, 2021.

\bibitem{zhao2017pyramid}
Hengshuang Zhao, Jianping Shi, Xiaojuan Qi, Xiaogang Wang, and Jiaya Jia.
\newblock Pyramid scene parsing network.
\newblock In {\em CVPR}, pages 2881--2890, 2017.

\bibitem{zheng2021rethinking}
Sixiao Zheng, Jiachen Lu, Hengshuang Zhao, Xiatian Zhu, Zekun Luo, Yabiao Wang,
  Yanwei Fu, Jianfeng Feng, Tao Xiang, Philip~HS Torr, et~al.
\newblock Rethinking semantic segmentation from a sequence-to-sequence
  perspective with transformers.
\newblock In {\em CVPR}, pages 6881--6890, 2021.

\bibitem{zhou2017scene}
Bolei Zhou, Hang Zhao, Xavier Puig, Sanja Fidler, Adela Barriuso, and Antonio
  Torralba.
\newblock Scene parsing through ade20k dataset.
\newblock In {\em CVPR}, pages 633--641, 2017.

\bibitem{zhu2021deformable}
Xizhou Zhu, Weijie Su, Lewei Lu, Bin Li, Xiaogang Wang, and Jifeng Dai.
\newblock Deformable detr: Deformable transformers for end-to-end object
  detection.
\newblock In {\em ICLR}, 2021.

\end{thebibliography}
}

\clearpage
\appendix
\onecolumn
\section*{Appendix}

We organize our supplementary material as follows. 
\begin{itemize}
    \item In Section~\ref{subsec:A1}, we investigate the potential for applying our DFPQ to instance segmentation.
    \item In Section~\ref{subsec:A2}, we show more comparisons on the visualized cross-attention maps for different positional queries.
    \item In Section~\ref{subsec:A3}, we show more qualitative results.
    \item In Section~\ref{subsec:A4}, we investigate the effect of different low-resolution features choices in HRCA.
    \item In Section~\ref{subsec:A5}, we investigate the effect of starting to employ DFPQ at midway training.
\end{itemize}

\section{Instance Segmentation with DFPQ}\label{subsec:A1}
To further demonstrate the flexibility of our method, we combine our DFPQ with Mask2former and compare with the baseline instance segmentation methods on COCO~\texttt{val2017}~\cite{lin2014microsoft} following the exact settings in~\cite{cheng2021masked}. The results are reported in Table~\ref{tab:coco_ins}. We observe that our DFPQ consistently improves Mask2former with R50 and Swin-B backbones by 0.3\% and 0.2\% AP with barely extra parameters and FLOPs. The results suggest the potential to extend our DFPQ to other segmentation scenarios. However, the improvements are not as impressive as in semantic segmentation. Instance segmentation is a very challenging task that requires grouping highly-entangled pixels into groups of instances and is more challenging than semantic segmentation or object detection as recognized by literature~\cite{li2022mask}. Previous work~\cite{carion2020end,he2017mask,xiao2018unified,li2022mask} provides the positional priors by integrating the instance segmentation with a heavy object detection head or branch in a two-stage top-down ~\cite{carion2020end,he2017mask,xiao2018unified} or a bottom-up style~\cite{li2022mask} framework. Compared to the literature that employs specific architecture designs to provide the positional priors, we conjecture that our DFPQ has an inferior representational capability that cannot fully encode the required positional priors for the challenging instance segmentation task. However, it is an interesting future direction to additionally encode the instance-level information into our DFPQ, \eg, encoding the bounding boxes or explicitly distinguishing the positional priors among the instance segments.

\begin{table}[h!]
\centering
\caption{Combine our DFPQ with Mask2former~\cite{cheng2021masked} and compare with the state-of-the-art instance segmentation methods on COCO \texttt{val}~\cite{zhou2017scene} with 133 categories. \#P and \#F indicate the number of parameters (M) and FLOPs (G).}
\resizebox{0.6\linewidth}{!}{%
\begin{tabular}{c|cc|cccccc}
Method & Backbone & Epochs & AP  & AP$^{\rm s}$ & AP$^{\rm m}$ & AP$^{\rm l}$ & \#P & \#F \\ \shline
SOLOv2~\cite{wang2020solov2} & R50 & 36 & 37.5 & 15.8 & 41.4 & 56.6 & 34 & -  \\
K-Net~\cite{zhang2021k} & R50 & 36 & 38.6 & 19.1 & 42.0 & 57.7 & 37 & - \\
HTC~\cite{chen2019hybrid} & R101 & 36 & 39.7 & 22.6 & 42.2 & 50.6 & 80 & 441 \\
Mask2former~\cite{cheng2021masked} & R50 & 36 & 42.4 & 22.1 & 45.4 & 64.3 & 44 & 225 \\
Mask2former + DFPQ & R50 & 36 & 42.7 & 21.9& 46.4 & 64.4  & 44 & 225 \\
\shline
Mask2former~\cite{cheng2021masked}& Swin-B & 36 & 47.2& 27.1 & 50.8 & 69.4 & 115 & 466 \\ 
Mask2former + DFPQ & Swin-B & 36 & 47.4 & 26.8 & 51.0 & 70.5 & 115 & 466  \\
\end{tabular}%
}
\vspace{-0.5em}
\label{tab:coco_ins}
\end{table}

\begin{figure*}[ht!]
  \centering
  \includegraphics[width=1.0\linewidth]{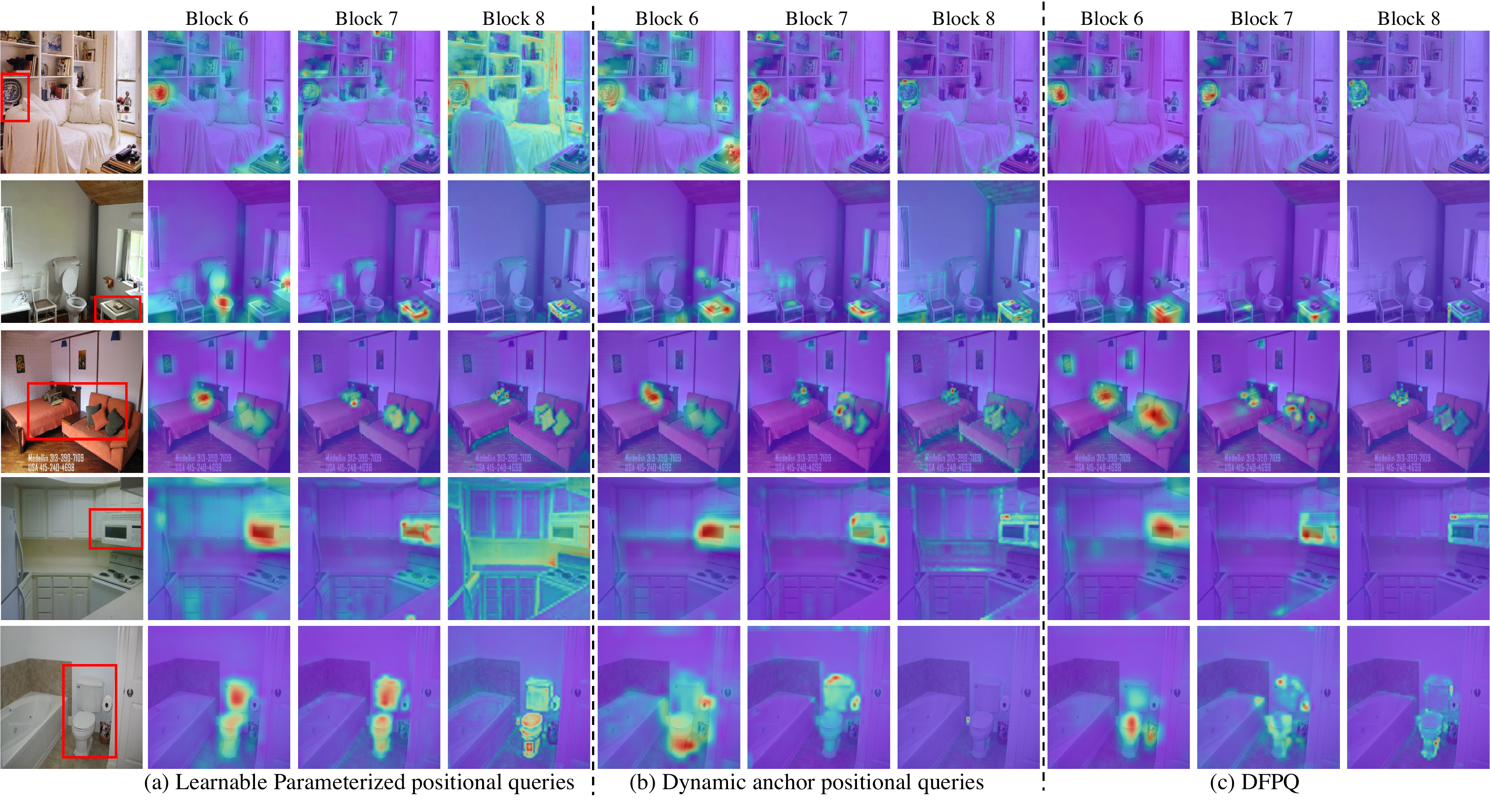}
  \vspace{-2em}
  \caption{Visualizations of the cross-attention maps for learnable positional queries (\cite{carion2020end,cheng2021masked}), dynamic anchor positional queries (alike \cite{liu2022dabdetr}) and our DFPQ. We show the visualizations for the normalized cross-attention maps in the last three decoder blocks and indicate the target segments in the red boxes. The cross-attention maps with the learnable positional queries and the dynamic anchor positional queries are often scattered without a clear focus and mix up different segments, while the cross-attention maps with DFPQ are more compact and consistent to reflect the target segments.}
  \vspace{-0.5em}
  \label{fig:more_attention}
\end{figure*}

\section{More Comparisons on Visualized Cross-attention Maps}\label{subsec:A2}

We have discussed and provided both quantitative and qualitative comparisons with other positional queries variants in Section 4.2 of the main paper. We show more comparisons on visualized cross-attention maps with different positional queries in Figure~\ref{fig:more_attention} and observe that the quality of the cross-attention maps for our DFPQ is clearly better than the learnable parameterized positional queries and dynamic anchor positional queries. Our DFPQ progressively refines the cross-attention maps, which become more accurate and compact in the deeper layers. Interestingly, we observe the cross-attention maps with DFPQ can end-to-end learn to localize the boundaries of the target segments in Block 8. 

\section{More Qualitative Results} \label{subsec:A3}
\begin{figure*}[ht!]
  \centering
  \includegraphics[width=1.0\linewidth]{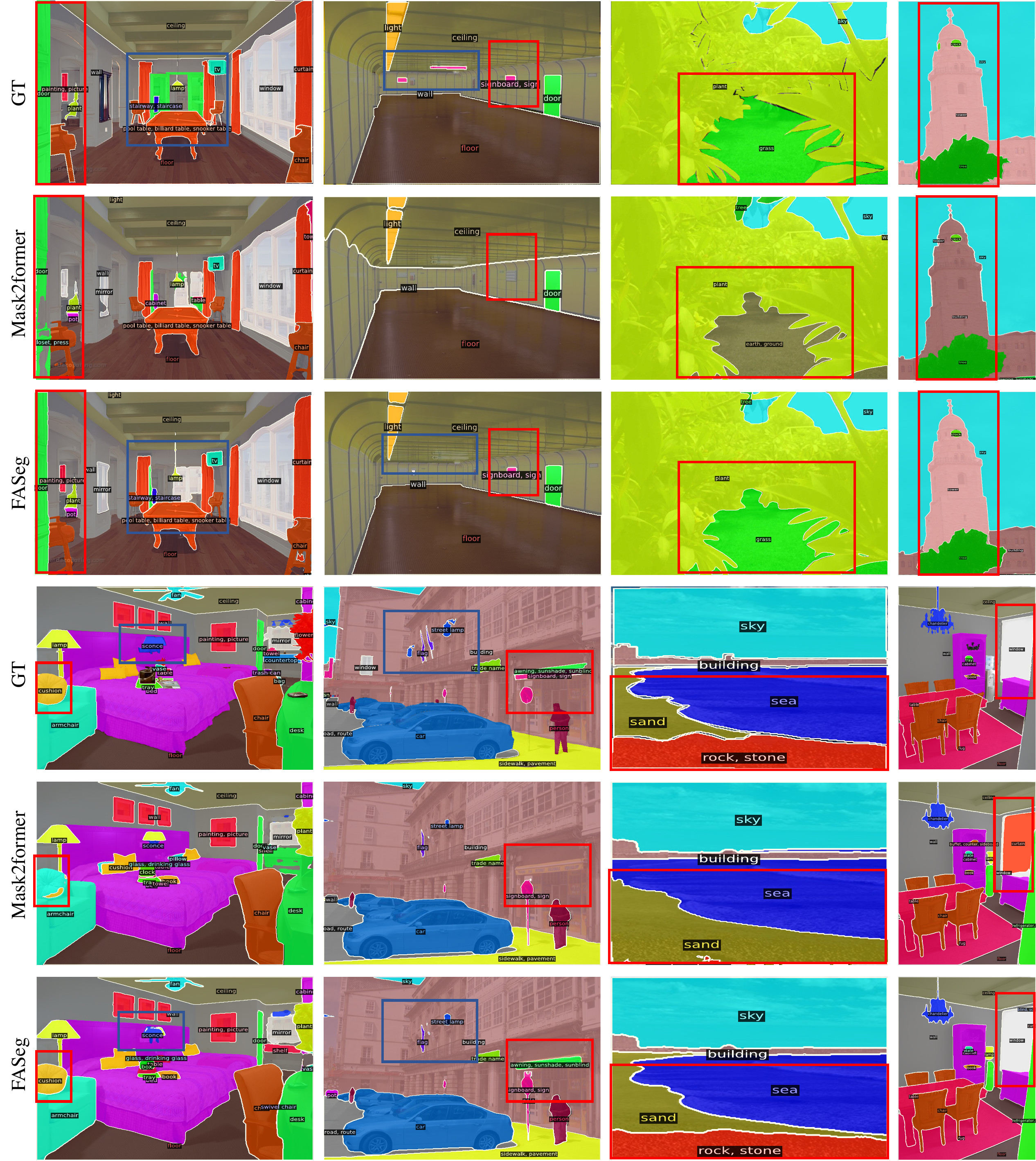}
  \vspace{-2em}
  \caption{Qualitative results on the ADE20K \texttt{val}~\cite{zhou2017scene}. Compared to Mask2former~\cite{cheng2021masked}, our FASeg predicts masks with finer details and yields more accurate predictions. The differences are highlighted with red boxes and the failure cases are highlighted with blue boxes. Best viewed in color.}
  \vspace{-0.5em}
  \label{fig:more_vis}
\end{figure*}

We visualize sample predictions of our FASeg model with Swin-L backbone and compare with Mask2former~\cite{cheng2021masked} on ADE20K \texttt{val} with multi-scale inference in Figure~\ref{fig:more_vis}. We observe that FASeg generates consistent predictions that align with the ground truth. We also present some failure cases in the blue boxes and find that the very small target segments still cannot be localized precisely. We thus take refining the small regions as future work.

\section{Effect of Low-resolution Feature scale Choices for HRCA}
\label{subsec:A4} In HRCA, we identify the most informative pixel positions within the low-resolution features, then map these positions to the high-resolution features and only perform cross-attention on these positions. We keep the number of attended pixels in the high-resolution features the same and investigate the effect of different low-resolution feature scales on ADE20k \texttt{val}. The results are reported in Table~\ref{tab:abl_resolution}. We observe that more coarse-grained features yield better results, where we conjecture that coarse-grained features contain more context information which helps locate the informative pixels. In this case, we use 1/32$\times$1/32 low-resolution features to select the informative pixels by default for the other experiments.

\section{Effect of Starting to Employ DFPQ at the Midway of Training} \label{subsec:A5}
By default, DFPQ is applied at the beginning of training. We experiment to start employing DFPQ from 20k, 40k, and 80k iterations of the total 160k training iterations for FASeg with Swin-B backbone on ADE20k \texttt{val}. The results are reported in Table~\ref{tab:abl_mid_dfpq}. We find that the performance fluctuates within 0.3\% mIoU, which suggests that our DFPQ is robust to the starting iteration. We also speculate that our DFPQ is robust to the early-stage training and will eventually learn reasonable positional information.

\begin{table}[tb!]
\centering
\caption{Effect of low-resolution features for HRCA on ADE20K \texttt{val}~\cite{zhou2017scene} with 150 categories.}
\vspace{-1em}
\resizebox{0.3\linewidth}{!}{%
\begin{tabular}{c|c}
Low-level features & mIoU s.s. (\%) \\ \shline
1/8$\times$1/8 & 47.9 \\
1/16$\times$1/16 &  48.0 \\
1/32$\times$1/32 & 48.3 \\
\end{tabular}%
}
\vspace{-1em}
\label{tab:abl_resolution}
\end{table}

\begin{table}[t!]
\centering
\caption{Effect of starting to employ DFPQ at midway training for FASeg with Swin-B Backbone on ADE20K \texttt{val}~\cite{zhou2017scene} with 150 categories.}
\vspace{-1em}
\resizebox{0.4\linewidth}{!}{%
\begin{tabular}{c|ccccc}
Starting iteration & 0 & 20k & 40k & 80k \\\hline
mIoU s.s.(\%) & 55.0 & 55.1 & 55.1 & 54.7 \\
\end{tabular}%
}
\vspace{-1em}
\label{tab:abl_mid_dfpq}
\end{table}

\end{document}